\documentclass[letterpaper]{article} 
\usepackage{aaai24}  
\usepackage{times}  
\usepackage{helvet}  
\usepackage{courier}  
\usepackage[hyphens]{url}  
\usepackage{graphicx} 
\usepackage{natbib}  
\usepackage{caption} 
\usepackage{algorithm}
\usepackage{algorithmic}

\usepackage{newfloat}
\usepackage{listings}
\DeclareCaptionStyle{ruled}{labelfont=normalfont,labelsep=colon,strut=off} 
\floatstyle{ruled}
\newfloat{listing}{tb}{lst}{}
\floatname{listing}{Listing}
\usepackage{subcaption}
\usepackage{amsmath}
\title{Small Dataset, Big Gains: Enhancing Reinforcement Learning by Offline Pre-Training with Model Based Augmentation}

\author {
    Girolamo Macaluso\textsuperscript{\rm 1},
    Alessandro Sestini\textsuperscript{\rm 2},
    Andrew D. ~Bagdanov\textsuperscript{\rm 1}
}
\affiliations {
    \textsuperscript{\rm 1}MICC - University of Florence\\
    \textsuperscript{\rm 2}SEED – Electronic Arts\\
    girolamo.macaluso@unifi.it, asestini@ea.com, andrew.bagdanov@unifi.it 
}

\begin{document}

\maketitle

\begin{abstract}
Offline reinforcement learning leverages pre-collected datasets of transitions to train policies. It can serve as effective initialization for online algorithms, enhancing sample efficiency and speeding up convergence. However, when such datasets are limited in size and quality, offline pre-training can produce sub-optimal policies and lead to degraded online reinforcement learning performance. In this paper we propose a model-based data augmentation strategy to maximize the benefits of offline reinforcement learning pre-training and reduce the scale of data needed to be effective. Our approach leverages a world model of the environment trained on the offline dataset to augment states during offline pre-training. We evaluate our approach on a variety of MuJoCo robotic tasks and our results show it can jump-start online fine-tuning and substantially reduce -- in some cases by an order of magnitude -- the required number of environment interactions.
\end{abstract}

\section{Introduction}

Effective policy learning with Reinforcement Learning (RL) often demands extensive interaction with the environment, a process that can be both expensive and potentially unsafe in real-world scenarios, such as robotics, logistics, and autonomous driving, where exploration with untrained policies is either costly or poses safety risks~\cite{levine2020offline}. In some cases, pre-collected experience datasets are available, offering a valuable resource.

Offline RL has emerged as a technique to leverage such dataset to train effective policies, eliminating the need for continuous interaction with the environment during training. In offline RL, no assumptions are made about the policy used to gather the data, known as the behavioral policy. The primary objective of offline RL is, indeed, to create a policy that outperforms the performance of the behavioral policy. However, this poses significant challenges. Enhancing the policy beyond the performance of the behavior policy that gathered the data involves estimating values for actions not encountered in the dataset. This introduces the risk of distributional shift, where the policy may face states or actions during deployment that differ significantly from those in the training data. To reduce this problems several offline RL approaches introduces constraints to the learned policy~\cite{levine2020offline}.

A promising approach to reduce sample complexity and training time of RL is taking advantage of pre-collected experience combining offline and online RL~\cite{nair2020awac, luo2023finetuning}, the first is used to create an initialization that jump starts the online training. However, we find that the size of the offline dataset significantly influences the effectiveness of offline pre-training, sometimes even slowing down the online process significantly. This is due to the overfitting that happens when combining offline RL and small datasets.

\begin{figure*}
    \centering
    \includegraphics[width=1.\textwidth]{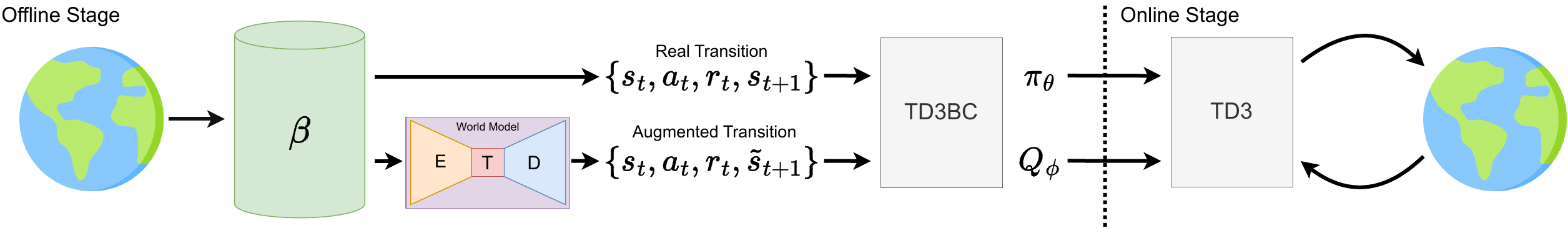}
    \caption{Our proposed training process: The offline dataset is used to train our world model (purple). We then use this model to augment half of the transitions sampled from the offline dataset $\beta$ in a batch of TD3BC training samples. The resulting actor $\pi_{\theta}$ and critic $Q_{\phi}$ are then used as initialization for the online learning phase with TD3.}
    \label{approach}
\end{figure*}

In this work, we introduce a model-based approach to enhance the offline RL pre-training stage (see Figure~\ref{approach}) that fully leverages potential of the offline dataset to speed up online training and significantly reduce the number of online environment interactions needed. Our proposed approach is based on a generative world model, trained end-to-end using the offline dataset, that can generate the next state given the actual state and the action. This capability is exploited during offline training to augment the transition used, creating a policy that is more informed about the environment. We conduct experimentation on a very small sample of the D4RL~\cite{d4rl} offline MuJoCo~\cite{mujoco} datasets to simulate a scarce data scenario. We also try to understand what is the impact of our augmentation on the learned initialization. We use Twin Delayed Deep Deterministic Policy Gradient (TD3) for the online fine tuning and its offline version Twin Delayed Deep Deterministic Policy Gradient with Behavioral Cloning (TD3BC) for the pre-training.

Our primary contribution is a framework designed to train policies, maximizing the information gained from small pre-collected experience datasets and enhancing sample efficiency. We begin by training a generative world model using the offline dataset, and subsequently, we leverage this model to augment the offline pre-training, enhancing the quality of the online initialization. This effectively accelerates the fine-tuning process, enabling the policy to achieve the same return value as a fully online-trained policy in significantly fewer iterations, sometimes even by an order of magnitude.    

\section{Related Work}
\label{sec:related}

In offline Reinforcement Learning (RL), we optimize a policy without relying on interactions with the environment, but rather by using a fixed dataset pre-collected from a behavioral policy~\cite{orl2,kumar2020conservative}. The task of offline RL is to train a policy that surpasses the performance of the behavioral policy, ``stiching" together part of different trajectories that contains good behaviors.
This framework has recently gained interest within the research community for its potential in scenarios where extensive interactions with the environment are costly or unsafe, such as in real-world applications like robotics, logistics, and autonomous driving~\cite{levine2020offline}. 
Another interesting use case of offline RL is to initialize an online RL training using a dataset of experiences. In this section we review work from the literature most related to our contributions.

\paragraph{Combining Offline and Online RL.}
The combination of offline and online RL techniques has emerged as a promising research direction. In works like~\cite{nair2020awac, nakamoto2023cal, luo2023finetuning}, offline RL has been used to train a policy from a precollected dataset of experiences that is then fine-tuned with online RL. These studies have investigated diverse strategies aimed at improving the performance gain of offline pretraining and mitigating the phenomenon known as \textit{policy collapse}, which causes a performance dip when shifting from offline to online training~\cite{luo2023finetuning}. This studies reveals that the dip may be caused by the initial overestimation of the offline trained critic for states-action pair unseen during training. 

These approaches propose measures such as reducing the underestimation during offline~\cite{nakamoto2023cal}, imposing a conservative improvement to the online stage~\cite{luo2023finetuning}, and weighting policy improvement with the advantage function~\cite{nair2020awac}.
 
Our work extends this research to scenarios with severely limited offline dataset, introducing a model-based augmentation technique that enhances offline pre-training, taking full advantage of the available data.
    
\paragraph{World Models.} World models (WM) have attracted great attention in prior RL research, demonstrating their capacity to capture environmental dynamics and elevate the performance of agents. 
A WM trained through random exploration has exhibited strong performance when used as a proxy for policy learning in simple environments~\cite{ha2018world}. In such cases, the WM excels at predicting future outcomes, effectively simulating the true environment.
The ``\emph{Dreamer}" framework~\cite{hafner2019drv1,hafner2020drv2} takes a more comprehensive approach, iteratively refining the WM. An agent is trained using the WM ``\textit{imagined}" trajectories and is then deployed to gather new experiences, that are used to refine the WM. Remarkably, this framework has demonstrated considerable success, even learning a policy from scratch to obtain diamonds in the popular video game \textit{Minecraft}, a very difficult task due to the gigantic state space and the sparsity of the reward~\cite{wu2023drv3}.
Our approach builds on this concept of integrating a WM into the training process, but instead of simulating trajectories we use the WM to augment the small available dataset during the offline training.
   
\section{Effective Offline Pre-Training with Scarce Data}
\label{sec:offlinepretrain}

Our method centers around the development of a generative WM utilizing the offline dataset to enhance offline RL pre-training. The primary objective is to generate a more informed actor-critic pair that serves as a superior initialization for subsequent online training and that also diminishes the need for extensive online interactions, resulting in a more efficient and effective RL training paradigm.

\subsection{World Model}
The core of our approach is the generative WM designed as a Variational AutoEncoder (VAE) with a transition model. This WM is responsible for encapsulating the environmental dynamics and providing the capability to predict the next state given the current state and action. The WM is composed of an variational encoder, a decoder, and the transition model.

\paragraph{Encoder-Decoder.} In our VAE setup, the encoder projects the environment state $s_t$ into a lower-dimensional latent space, parameterizing the mean $\mu_{s_t}$ and log variance $\log\sigma_{s_t}^2$ of a Gaussian distribution within this space. The current state's latent representation $z_t$ is sampled from this distribution, introducing crucial stochasticity for capturing diverse environment dynamics. Our experiments will demonstrate that this stochasticity can prove beneficial, even in environments that are deterministic. This sampling allows our WM to generate different possible next states, for the same state-action pair. Conversely, the decoder reconstructs the latent state back into the original state space. In our approach, the decoder is employed to project the next latent space $\widetilde{z}_{t+1}$, produced by the transition model, to the state space $\widetilde{s}_{t+1}$.

\paragraph{Transition Model.} The transition model plays the role of predicting the changes that occur within the latent space due to the execution of an action. The ability to model these changes allows our system to predict future states and thus simulate forward in time. The transition model takes as input the latent representation of the current state $z_t$ and the action $a_t$ to be performed. Utilizing this information, the transition model predicts the change $\Delta(s_t,a_t)$ in the latent space that would occur due to the action $a_t$. This is then added back to the current latent state representation $z_t$ to obtain the next latent state $\widetilde{z}_{t+1}$, represented as $\widetilde{z}_{t+1} = z_t + \Delta(s_t,a_t)$. This provides an estimate of the subsequent state's latent representation in the environment given the current state and action. We designed the transition model to predict the changes within the latent space rather than modeling the next latent space directly, making it a model of the variation induced by an action within a specified latent space.

    \begin{figure}[t]
    \centering
    \includegraphics[width=0.5\textwidth]{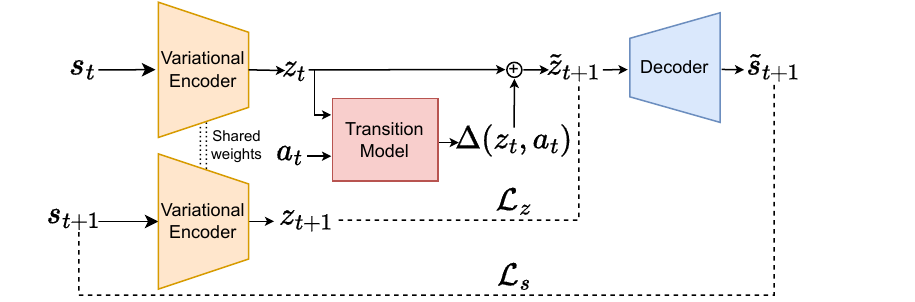}
    \caption{Our world model training procedure. The encoder (orange) takes the state $s_t$ and outputs mean and variance of a Gaussian used to sample the latent representation $z_t$. Then, the transition model (red) takes $z_t$ and action $a_t$ and generates the change in the latent space $\Delta(s_t,a_t)$ caused by the action $a_t$ in the latent state $z_t$. This is added back to the latent state to reconstruct the latent representation of the next state $\widetilde{z}_{t+1}$. The decoder (blue) then reconstructs the next state $\widetilde{s}_{t+1}$ starting from $\widetilde{z}_{t+1}$. The real next state $s_{t+1}$ is forwarded into the variational encoder obtaining $z_{t+1}$ that is used as target in the latent space reconstruction loss $\mathcal{L}_z$. The state space reconstruction loss in the state space $\mathcal{L}_s$ is instead computed between $s_{t+1}$ and $\widetilde{s}_{t+1}$.}
    
    \label{wmdiagram}
\end{figure}
    
\paragraph{Training Procedure.} Our WM is trained end-to-end on the offline dataset, with the objective of enabling the model to generate the next state given the current state and action.  Our model optimizes the \mbox{Evidence Lower Bound (ELBO)} loss along with two additional reconstruction losses, illustrated in Figure~\ref{wmdiagram}. 
\begin{align*}
    \mathcal{L}_\textnormal{wm}  & =
    \underbrace{\textnormal{MSE}(\widetilde{s}_{t},s_{t})+ \textnormal{KL}(\mathcal{N}(0,I) \, ||  \,\mathcal{N}(\mu_{s_{t}},\sigma^2_{s_{t}}))}_{\mathcal{L}_{\textnormal {ELBO}}}+\\
    & +\underbrace{\textnormal{MSE}(\widetilde{s}_{t+1},s_{t+1})}_{\mathcal{L}_s}+\underbrace{\textnormal{MSE}(\widetilde{z}_{t+1},z_{t+1})}_{\mathcal{L}_z}.
\end{align*}
The ELBO loss, a standard component in VAEs, plays a pivotal role in ensuring the meaningfulness of the learned latent representation. Its contribution lies in shaping a compact and dense latent space. Given that our WM optimizes this loss, it can effectively function as a variational autoencoder when utilizing only the encoder and decoder components.
The training of the transition model incorporates two reconstruction losses: one for the latent space $\mathcal{L}_z$ and another for the state space $\mathcal{L}_s$. The latent space loss aims to maintain coherence between the predicted latent representation and the one sampled by the encoder using the actual next state. Simultaneously, the state space loss ensures the model's ability in reconstructing the next state within the state space. Both reconstruction losses are calculated using the Mean Squared Error (MSE) metric.

\begin{figure*}[ht]
    \centering
    \begin{center}
        \includegraphics[width=0.4\textwidth]{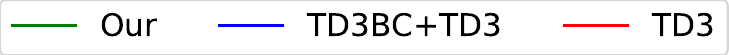}
    \end{center}

    \raisebox{0.15in}{\rotatebox[origin=t]{90}{\tiny Random~~~~~~~~~~~~~~~~~~~~~~~~~~~~~~~~~~~~~Medium-Replay~~~~~~~~~~~~~~~~~~~~~~~~~~~~~~~~~~Medium-Expert}} 
    \begin{subfigure}[c]{0.24\textwidth}
        \includegraphics[width=\textwidth]{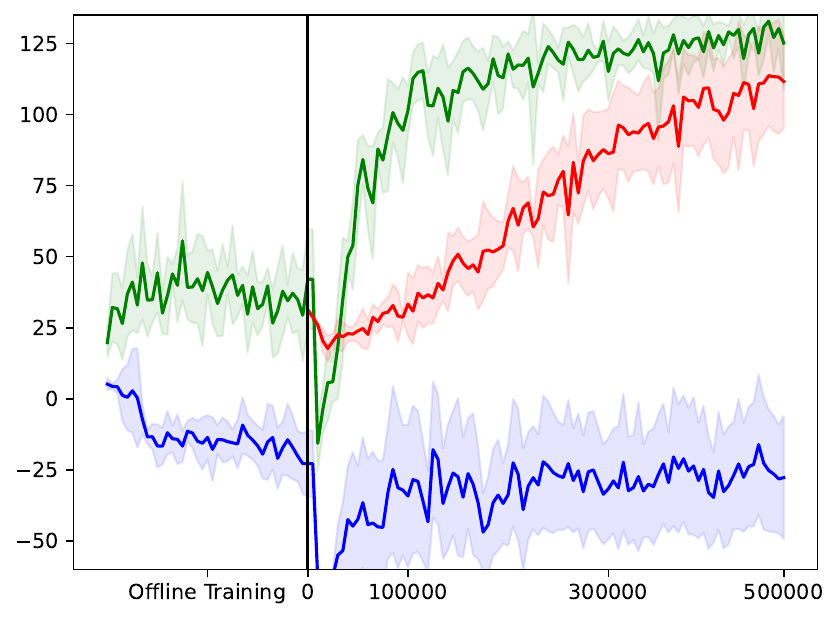}
        \includegraphics[width=\textwidth]{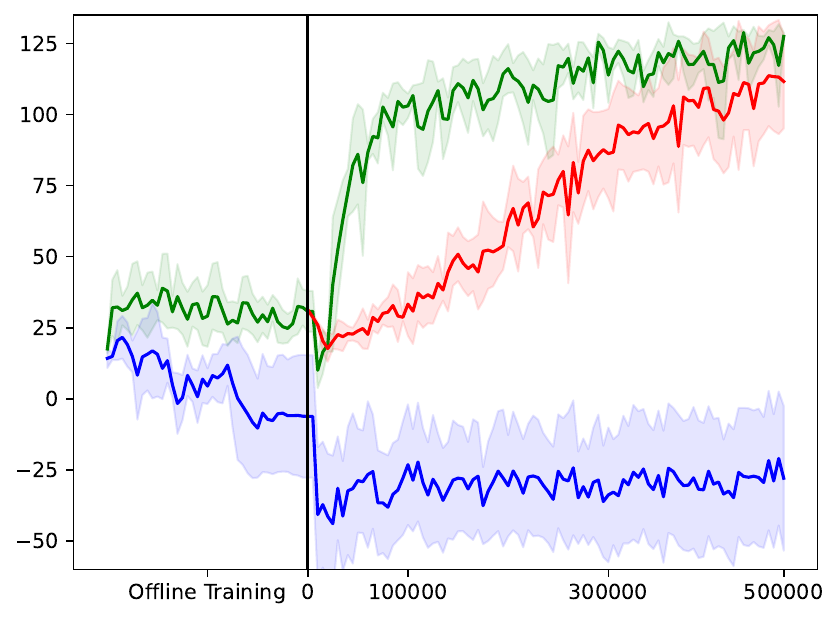}
        \includegraphics[width=\textwidth]{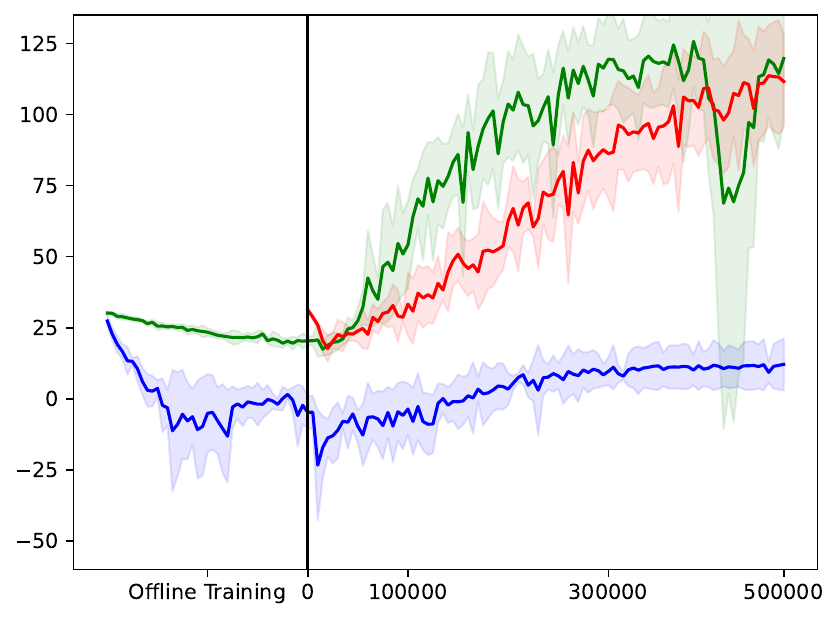}
        \caption{Ant}            
    \end{subfigure}
    \begin{subfigure}[c]{0.24\textwidth}
        \includegraphics[width=\textwidth]{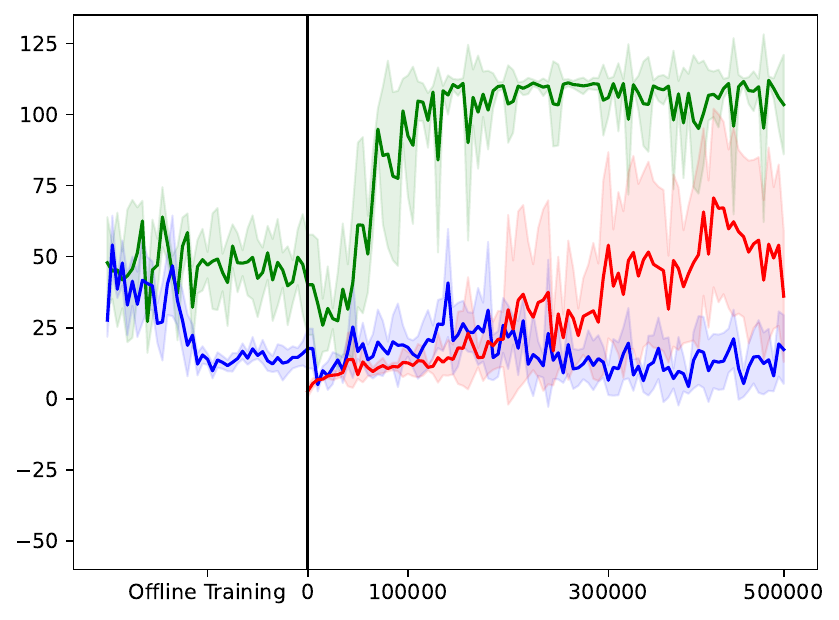}
        \includegraphics[width=\textwidth]{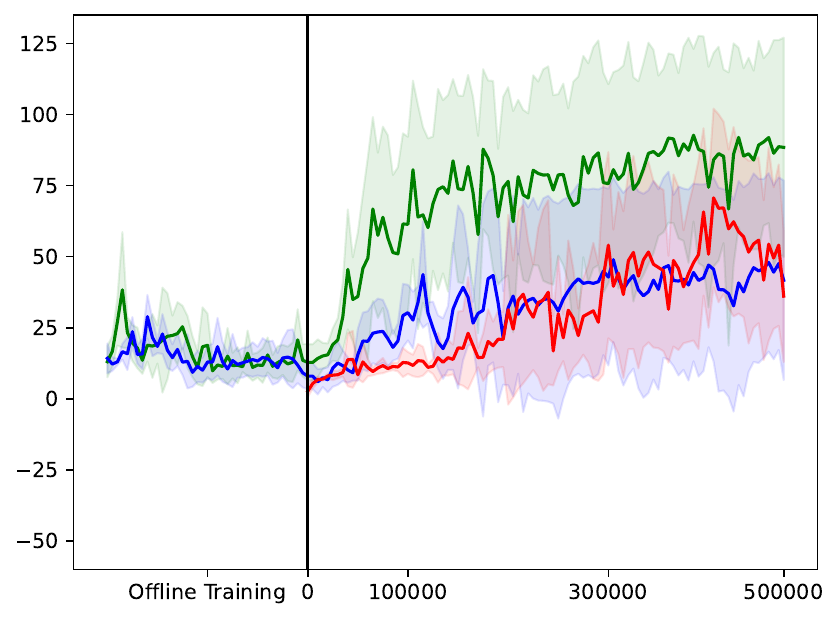}
        \includegraphics[width=\textwidth]{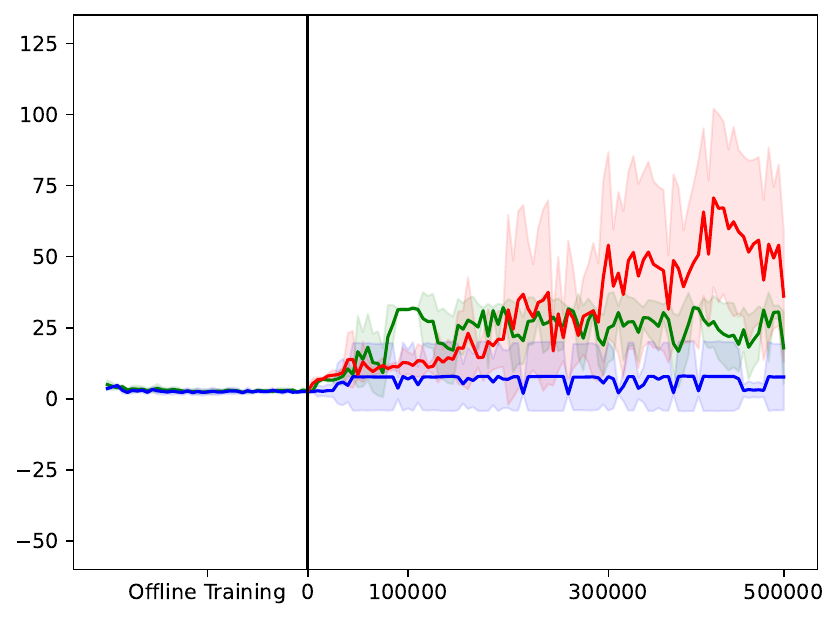}
        \caption{Hopper}            
        \end{subfigure}
    \begin{subfigure}[c]{0.24\textwidth}
        \includegraphics[width=\textwidth]{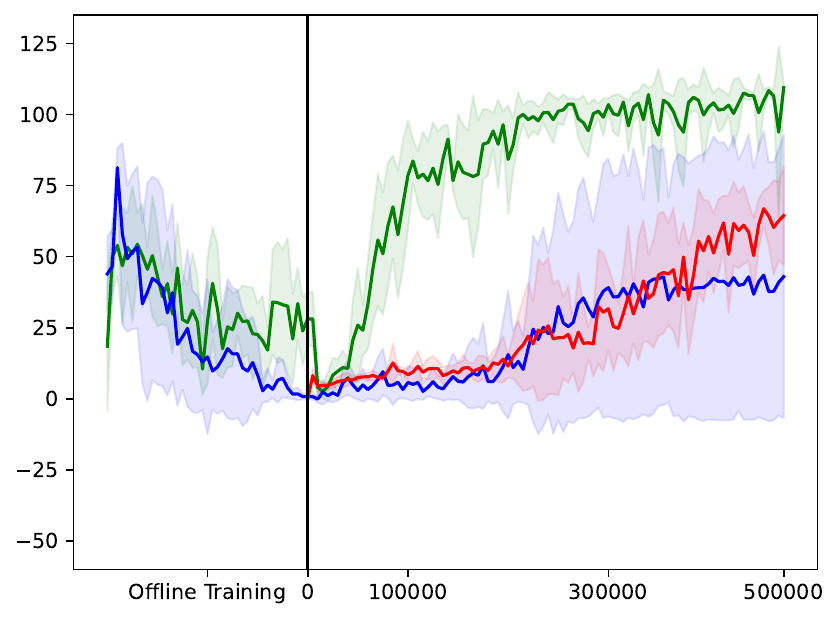}
        \includegraphics[width=\textwidth]{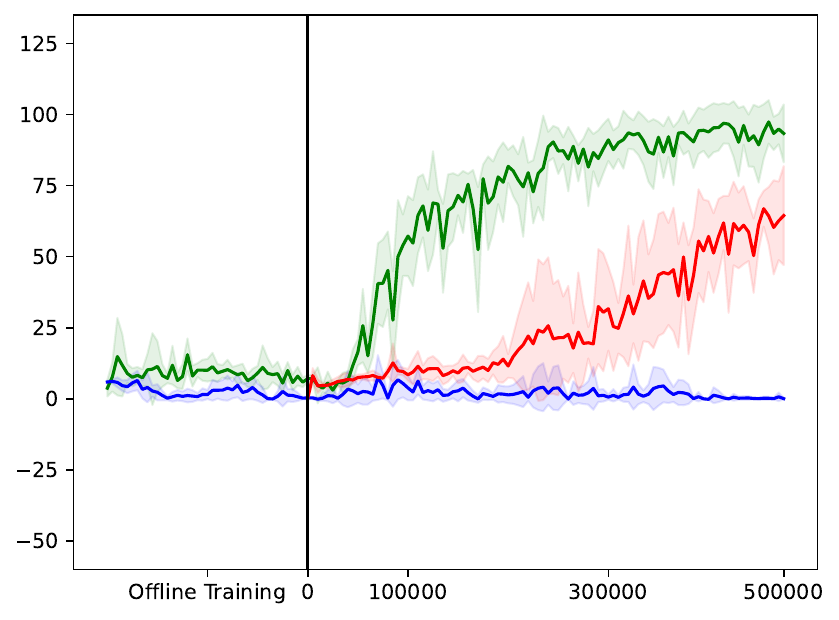}
        \includegraphics[width=\textwidth]{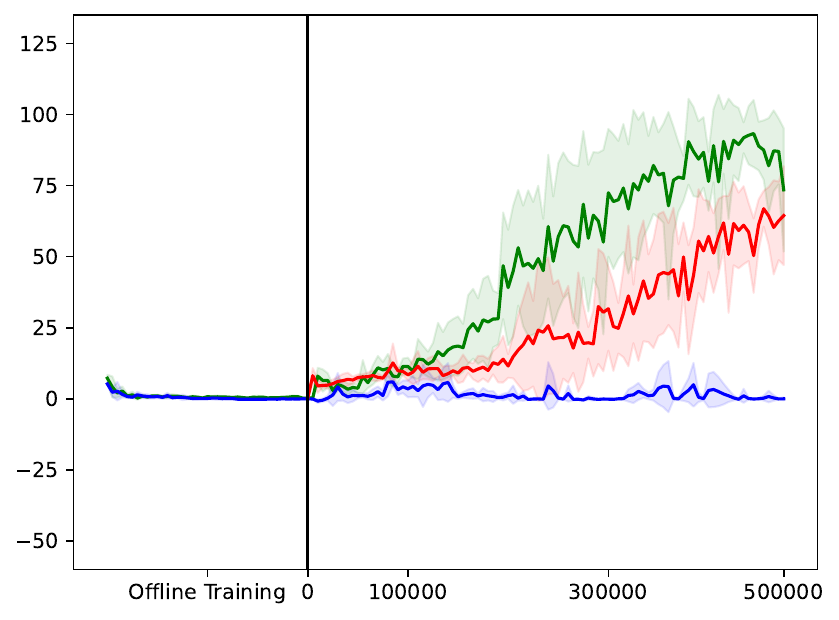}
        \caption{Walker}            
    \end{subfigure}
    \begin{subfigure}[c]{0.24\textwidth}
        \includegraphics[width=\textwidth]{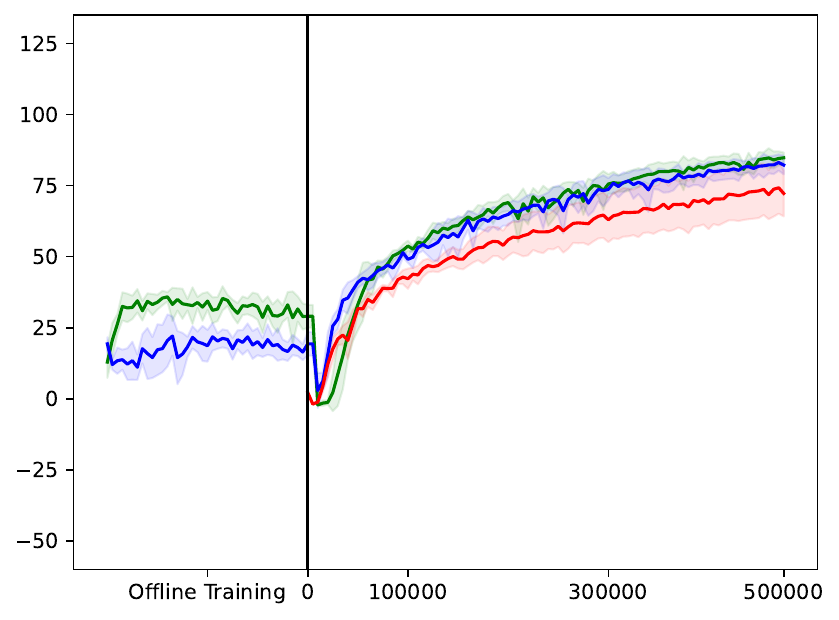}
        \includegraphics[width=\textwidth]{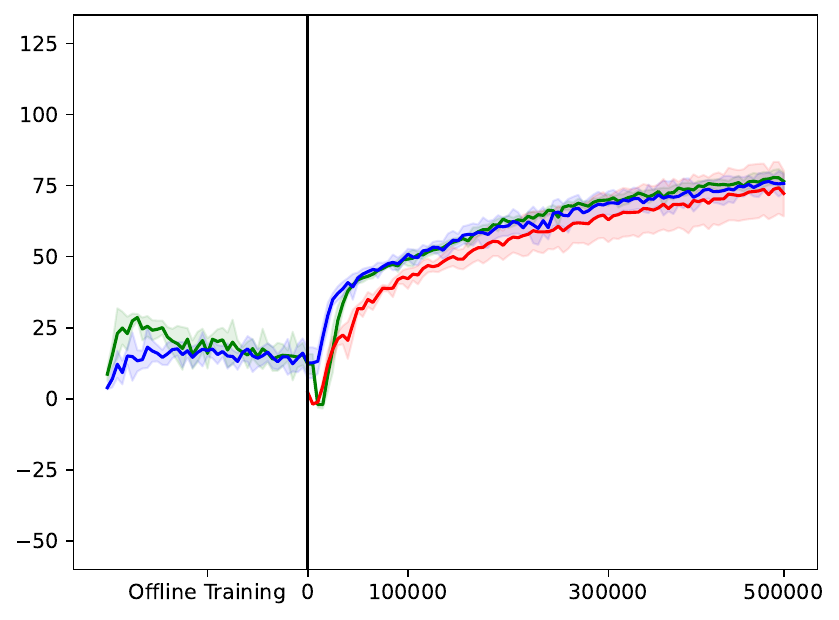}
        \includegraphics[width=\textwidth]{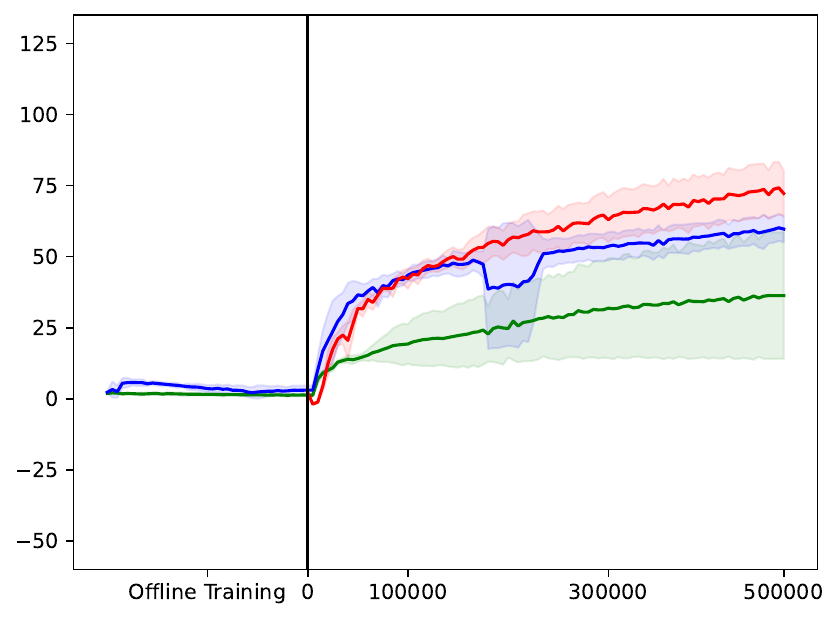}
        \caption{Halfcheetah}            
    \end{subfigure}
    \caption{Performance comparison between our approach (green line), TD3BC followed by TD3 (blue line), and full online training with TD3 (red line), measured in terms of normalized score. The segment preceding the black vertical line represents the scores achieved during offline training, while the subsequent segment reflects the transition to online training.}
    \label{results}
\end{figure*}
\subsection{Offline Pre-Training}

We perform the offline training with Twin Delayed Deep Deterministic Policy Gradient with Behavior Cloning (TD3BC)~\cite{fujimoto2021minimalist}, an actor-critic off-policy offline RL algorithm.
During the training a part of the transition within the batch sampled from the offline dataset is augmented with the use of our WM.
The augmentation consists in substituting the next state of 50\% of the transitions sampled in a training batch of TD3BC with one generated by our WM. The sampled transition tuple $(s_t, a_t, r_t, s_{t+1})$ becomes $(s_t, a_t, r_t, \widetilde{s}_{t+1})$, with $\widetilde{s}_{t+1}$ the state predicted by the WM using the current state $s_t$ and the action $a_t$. Augmented next states are not stored; each time a transition is sampled from the replay buffer, it can be dynamically augmented with a newly generated next state. 

Our augmentation technique has an impact on the computation of target Q-values, which plays a central role in the temporal difference learning of the critic of TD3BC~\cite{fujimoto2018addressing}.    
In the conventional approach, the target Q-values rely exclusively on the next states present in the offline dataset. This can lead to overfitting of the Q-function if the dataset is particularly small. This overfitting manifests itself as narrow peaks in the state-action value space, each corresponding to states observed in the dataset. Overfitting the Q-function can be detrimental to the overall performance of the offline RL algorithm, as it may struggle to generalize effectively to unseen or underrepresented states in the environment.  

By introducing the augmented next state generated by our WM into the computation of target Q-values, we aim to mitigate this risk and smooth out the narrow peaks by enriching the state-action value space with a more diverse set of states. This regularizing effect promotes a broader and more robust representation of environment dynamics and makes the TD3BC algorithm less sensitive to parts of the environment that may not be adequately represented in the offline dataset.

Intuitively this augmentation strategy serves as a guided exploration of the state space that aligns with the WM's interpretation of the environmental transition dynamics. It leverages the WM's predictive capabilities to envision different potential future outcomes and effectively expand the states considered during offline training.

We decided to employ our WM for generating predictions only one step into the future. This cautious decision is in line with the constrained size of the dataset, intending to find a balance that utilizes the WM's predictive capabilities while minimizing the risk of accumulating errors caused by data scarcity.
    
\subsection{Online Fine-tuning}
Our offline trained actor and critic serves as initialization for the online RL phase with Twin Delayed Deep Deterministic Policy Gradient (TD3)~\cite{fujimoto2018addressing}. 
We maintain the TD3BC framework's setup for the offline stage, specifically retaining state normalization. Upon transitioning to TD3, we continue normalization to effectively leverage the actor and critic. To perform the normalization we utilize the statistics from the offline dataset. The transition from TD3BC to TD3 is straightforward, involving the removal of the behavioral cloning term from the actor loss and the initialization of the online replay buffer with transitions sampled using the actions chosen by the pretrained actor.

The offline initialization aims to accelerate training by reducing the need for exhaustive online interaction with the environment. Offline training with data augmentation leads to a more informed actor-critic pair that jump-starts the online training by offering more reliable insight into state-action pairs. In the following section, we present an ablation study investigating the influence of the offline initialization of only actor or critic on the online training performance, aiming to evaluate the individual roles of each component.


\section{Experimental Results}

\begin{figure*}[ht]
\setlength{\tabcolsep}{2pt}
    \begin{center}
        \includegraphics[width=0.4\textwidth]{img/legend.pdf}
    \end{center}
    \begin{tabular}{ccc}
        \includegraphics[width=0.33\textwidth]{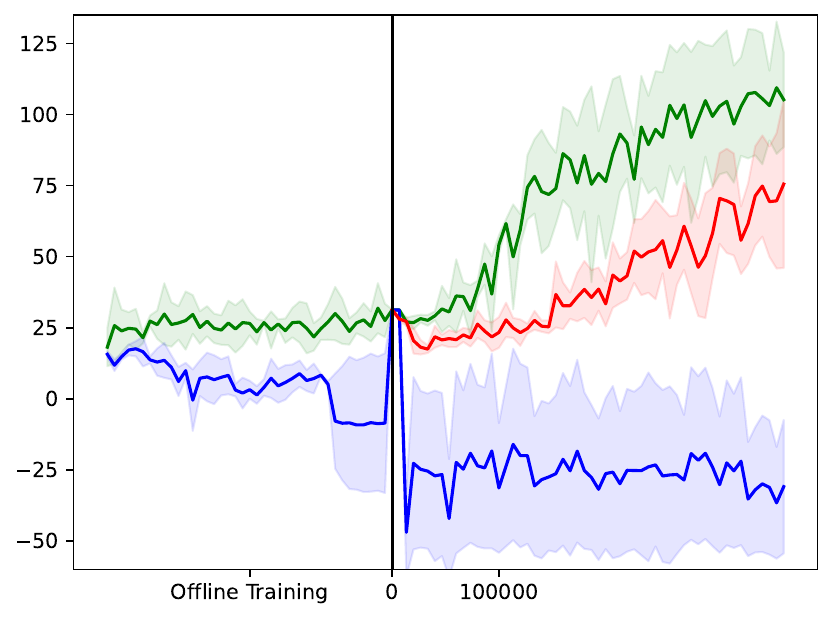} &
        \includegraphics[width=0.33\textwidth]{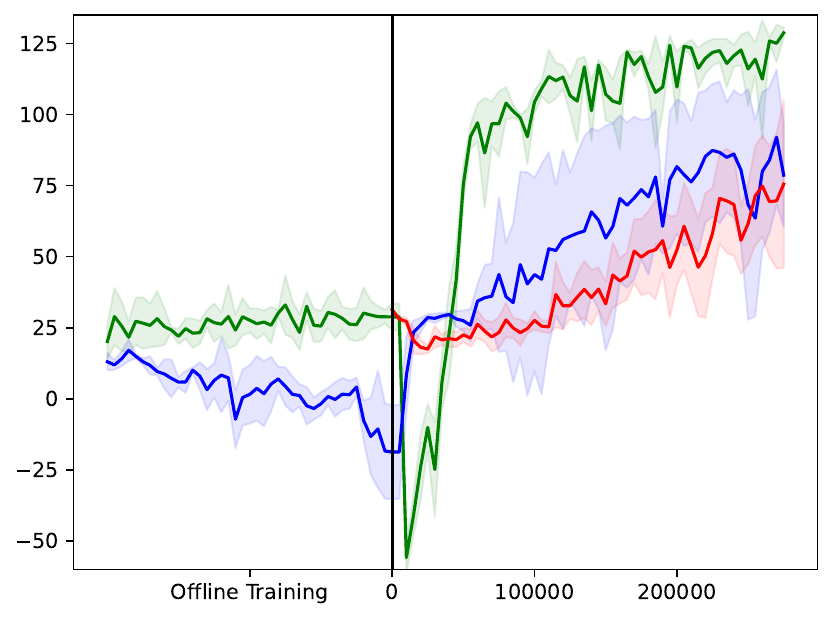} &
        \includegraphics[width=0.33\textwidth]{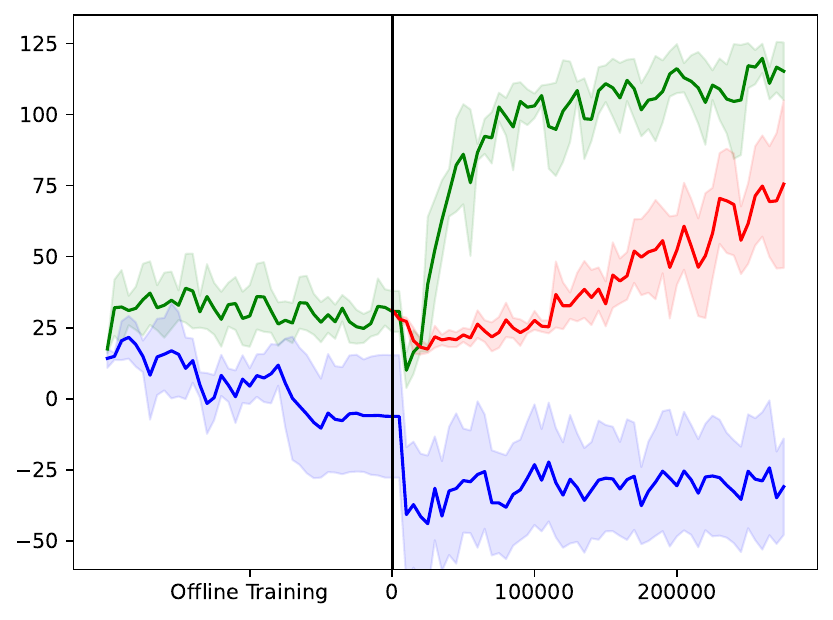} \\
        \scriptsize (a) \scriptsize Pre-training only actor. & \scriptsize (b) Pre-training only critic. & \scriptsize (c) Pre-training actor and critic.
    \end{tabular}
    \caption{Ablation on offline pre-training of online actor and critic. These results demonstrate that pre-training of both actor and critic contribute to the accelerated online learning.}
    \label{fig:ablations}
\end{figure*}

\label{sec:result}
In this section we report on experiments that demonstrate the ability of our approach to perform data augmentation and improve pre-training, which in turn enables more effective and efficient online fine-tuning.

\subsection{Experiments}
The architecture of our encoder, decoder, and transition model is a four layer Multi Layer Perceptron (MLP), with hidden layer size 512, and Rectified Linear Unit (ReLU) activations. The architecture of both policy and critic is the same used in the original TD3BC paper~\cite{fujimoto2021minimalist}, a three layer MLP with ReLU activations and hidden layer size of 256. 

We conducted experiments on four MuJoCo locomotion tasks: ant, hopper, walker and halfcheetah~\cite{mujoco}. Our experiments will demonstrate that our approach is beneficial, even in this deterministic environments. For each task we used the \textit{medium-expert}, \textit{medium-replay}, and \textit{random} datasets from D4RL~\cite{d4rl}. We sampled 10,000 transitions from each of these datasets, which represent only a tiny fraction of the original 2 million transitions for \textit{medium-expert}, 200,000 for \textit{medium-replay}, and 1 million for \textit{random}. These samples of the original transitions are used as our offline datasets in order to simulate a scarce data scenario. These datasets are leveraged to train our WM and for the offline RL training with TD3BC.

The offline stage is performed using our data augmentations for 200,000 iterations. The final online fine-tuning stage, initialized with the offline trained actor and critic, is then performed for another 500,000 iterations. 

We compare our approach with two baselines:
\begin{itemize}
    \item \textbf{TD3BC+TD3}, in which we use vanilla TD3BC for offline training and then use the learned policy and critic to initialize online learning with TD3 (initializing the replay buffer using the actions from the actor, as in our approach).

    \item \textbf{Fully Online TD3}, in which we train the policy and the critic from scratch using the off-policy online RL algorithm TD3. 
\end{itemize}
Results are evaluated in terms of normalized score~\cite{d4rl} which is zero when the evaluated policy has the performance of a random policy, 100 when it has the same performance of a policy trained online with the Soft Actor Critic (SAC) algorithm~\cite{haarnoja2018soft}, and over 100 if surpasses that performance. Evaluation is performed each 5,000 training iterations. We also show an ablation study on the role of the actor and the critic in jump starting the online training and an intuition of why our augmentation is effective.

\subsection{Comparative Performance Evaluation}

Figure~\ref{results} illustrates the results of our experiments on all environments. We observe a consistent pattern across nearly all configurations, when transitioning from offline to online there is evidence of \textit{policy collapse} in which the performance drops significantly~\cite{luo2023finetuning}. As observed by~\citet{luo2023finetuning}, when the data in offline dataset is more diverse, e.g. in \textit{medium-replay} and \textit{random} datasets, this performance dip is reduced. 

These results show our augmented offline initialization significantly improves the online training performance after the initial dip. This is due to the more informed actor-critic pair that better mitigates overestimation when encountering states not seen during offline training. This improvement in learning speed reduces the number of online interactions needed to learn a performant policy, as well as reducing sample complexity and the risks linked to initial exploration in real world environments. 

Figure~\ref{results} also illustrates the potential drawbacks of combining offline and online RL with limited data when not utilizing our augmentation. In all datasets and environments, employing vanilla TD3BC for policy initialization, in the best-case scenario, maintains similar performance to fully online training. However, it can also lead to a complete failure of the online training, showcasing the challenges associated with scarce data. This becomes evident across multiple configuration, including those from the ant environment, as well as in walker with \textit{random} and \textit{medium-replay} datasets, and hopper with \textit{random} and \textit{medium-expert} datasets. 

The performance of our technique is linked to the quality of the dataset available: when a dataset with expert or medium transitions are available the performance improves more consistently. Note how in the \textit{ant}, \textit{hopper}, and \textit{walker} environments we are able to achieve the same performance of SAC (i.e. a normalized score of 100) using only 100,000 online iterations when at least \textit{medium} quality transitions are included in the dataset. When the dataset, in addition to being small, is composed of randomly collected transitions the task become challenging. However we still note a small improvement in \textit{ant} and \textit{walker} environments. In the \textit{halfcheetah} environment, however, the use of our augmentation does not provide any notable improvement, and is even \textit{detrimental} when combined with a \textit{random} dataset.
\begin{figure*}[ht]
\setlength{\tabcolsep}{2pt}
    \begin{tabular}{cccc}
        \includegraphics[width=0.24\textwidth]{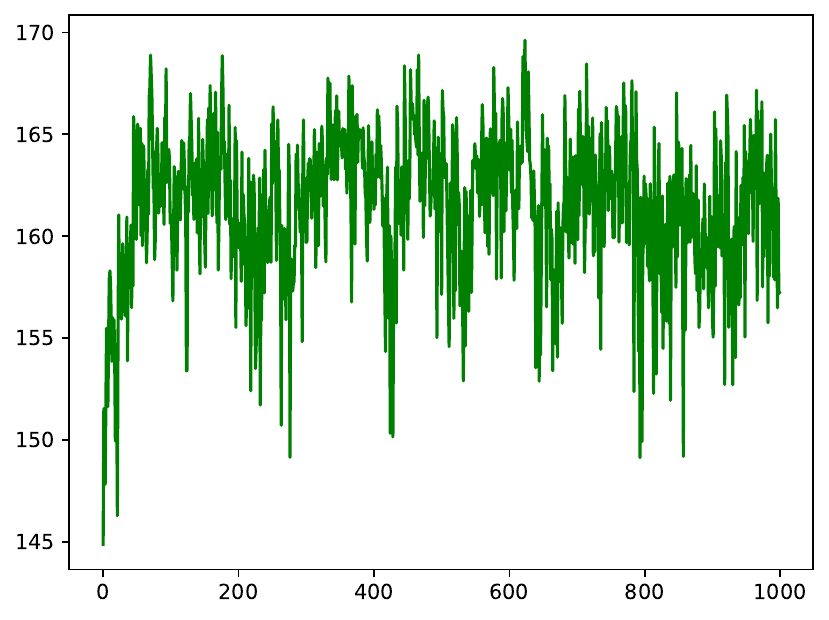} &
        \includegraphics[width=0.24\textwidth]{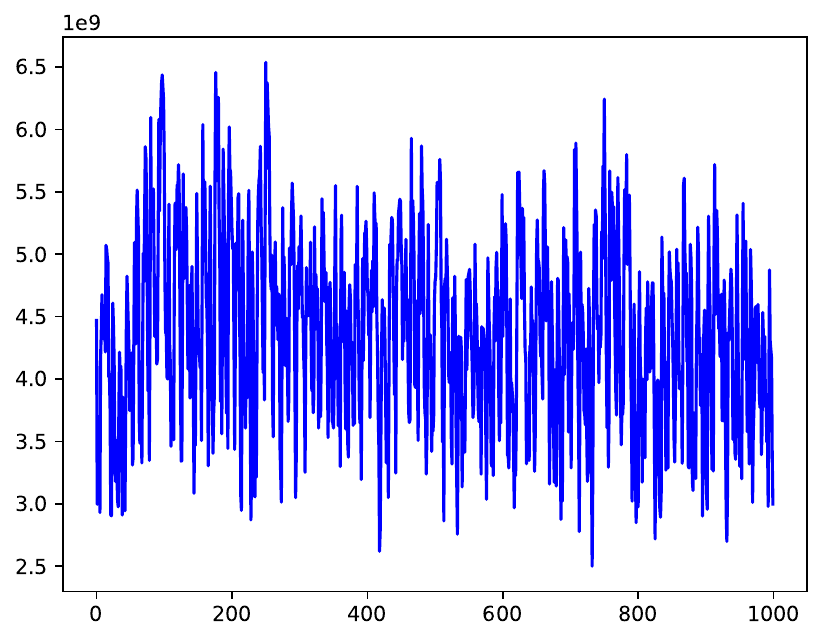} &
        \includegraphics[width=0.24\textwidth]{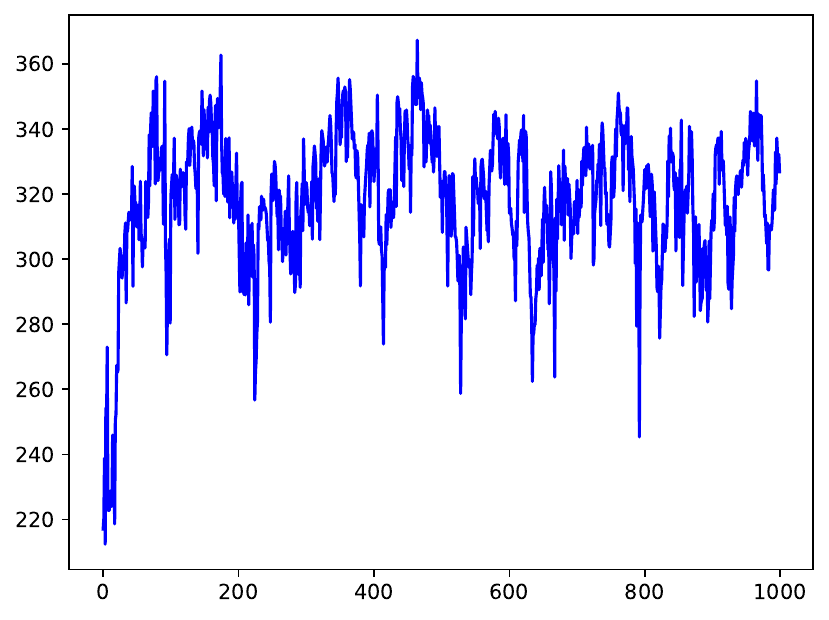} &
        \includegraphics[width=0.24\textwidth]{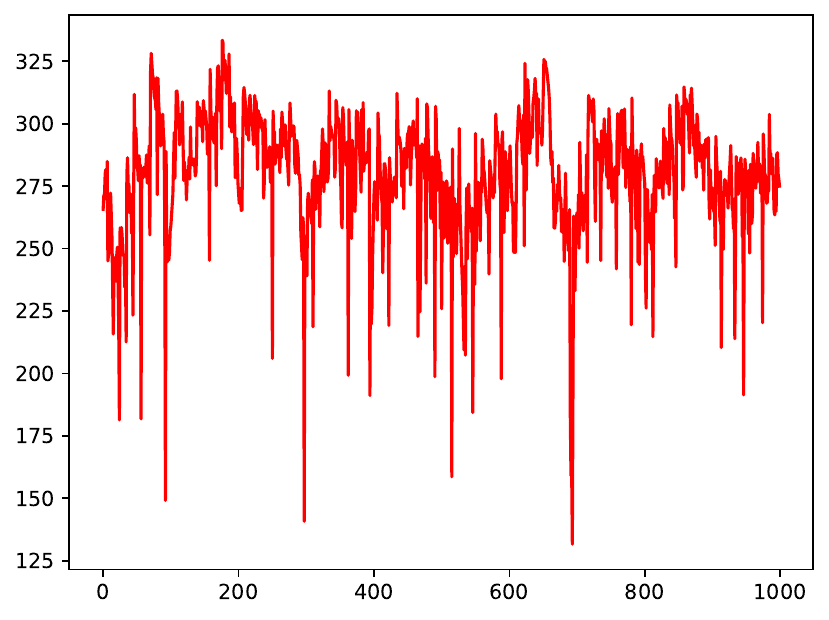} \\
        \scriptsize (a) \scriptsize 10K transitions w/ aug. & \scriptsize (b) 10K transitions w/o aug. & \scriptsize (c) 50K transitions w/o aug. & \scriptsize  (d) Online trained policy.
    \end{tabular}
    \caption{Q-values of critic networks over the same single episode for (a) our approach with 10,000 transitions, (b) pre-training on 10,000 transitions \textit{without} augmentations, (c) pre-training with TD3BC on 50,000 transitions without augmentations, and (d) online training with TD3.} 
    
    \label{fig:why-effective}
\end{figure*}

\subsection{Ablations}
\label{sec:ablation}
We performed ablations on the offline initialization of actor and critic to investigate if and how both contribute to the improved online learning performance observed in Figure~\ref{results}. In Figure~\ref{fig:ablations} we give the comparison of using both the offline pre-trained actor and critic on the online initialization~\ref{fig:ablations}(c), only critic~\ref{fig:ablations}(b), and only actor~\ref{fig:ablations}(a). This experiment was conducted on the \textit{ant medium-replay} dataset using an offline dataset consisting of only 10,000 transitions.  

Using only the offline trained actor still yields a consistent boost in performance compared to the baselines. It is noteworthy that in this scenario there is no performance dip. This is because policy collapse is linked to an initial overestimation of the critic when encounters state-action pairs out of the training distribution~\cite{luo2023finetuning}. This type of initialization may be useful in contexts in which is unsafe to have a sharp decrease in performance.

On the other hand, when initializing the critic we observe a substantial improvement following a significant dip in performance. This could be attributed to the inadequate initialization of the replay buffer, stemming from the use of a random actor during the initial online exploration. Additionally, the absence of the actor's balancing effect exacerbates the initial overestimation, contributing to the observed dip in performance. When considering the training using both actor and critic we observe a combination of the two behavior, with better performance than using only the actor with a much smaller dip.   

Without our data augmentation, using only the offline trained actor is detrimental to the training process and causes online training to fail completely. This highlights the critical role of the pre-trained actor when our augmentation is not applied. In such cases, the quality of the initialization becomes pivotal, as a poor starting actor can lead to a complete failure in the online fine-tuning, even if a pre-trained critic is used. On the other hand, not initializing the actor helps to avoid failure of the online fine tuning. However, the offline pre-training is still less beneficial then use our augmentation, as the results are marginally above to the one with both policy and actor random initialized, indicated in red.

\subsection{Why Pre-training with Augmentation is Effective}

To gain deeper insight into the impact of data augmentation with our generative WM, we conducted a comparative analysis of the critic value. Specifically, we compare the critic value after training with our augmentation using 10,000 transitions from the ant \textit{medium-replay} dataset. This evaluation is compared against a critic trained using 10,000 transitions on the same task without augmentation, a critic trained with same setup as the latter but with 50,000 transitions and a critic trained only online for 1 million iterations.

In Figure~\ref{fig:why-effective} we plot the results of this evaluation. In these plots, we show the value of the critic on the state-action pairs from the same episode sampled randomly from the \textit{expert} D4RL dataset. 
We see in Figure~\ref{fig:why-effective}(b) that the value of the critic trained without augmentation on 10,000 transitions falls drastically out of the environment's reward scale. This is likely due to the overfitting of the critic on the small offline dataset. However, in Figure~\ref{fig:why-effective}(c), as we increase the dimension of the offline dataset, we observe a more aligned critic resembling the one learned during the online stage, shown in Figure~\ref{fig:why-effective}(d).   

In contrast, our augmentation yields a more conservative critic that closely aligns with the distribution of the online trained critic, as illustrated in Figure~\ref{fig:why-effective}(a). This suggests that our approach helps in maintaining a more reliable and cautious critic estimation, avoiding extreme values that could lead to overestimation issues.
We believe that our augmentation process, involving the substitution of the evaluation state for the target critic during temporal difference learning with one generated by our WM, is simulating more exploration of the environment via the WM. This mechanism appears to contribute to a reduction of the overfitting.

\section{Conclusion}
\label{sec:conclusion}

In this paper we proposed an approach to effectively leverage small datasets to reduce sample complexity of reinforcement learning, through offline reinforcement learning initialization. Our approach is based on a generative world model, trained on the offline dataset, that is able to predict state transitions. We propose an augmentation performed during the offline training, based on the world model. Our experimental results shows that conventional offline-to-online training, with limited dataset and without our augmentation, yield ineffective or even detrimental results. On the other hand, our approach offers a solution that maximizes the utility of the small offline dataset, successfully training a meaningful initialization that is able to speed up the online training. This approach holds promise for improving the practical applicability of reinforcement learning in data-limited scenarios.

As future and ongoing work we are adapting this technique to more challenging environments like the adroit manipulation tasks~\cite{rajeswaran2017learning} in order to better assess the impact of data augmentation. We are also exploring the applicability and the impact of this technique on different offline and online reinforcement learning algorithms.

\bibliography{SmallDatasetBigGains}

\end{document}